\documentclass[runningheads]{llncs}
\usepackage{graphicx}
\usepackage{comment}
\usepackage{amsmath,amssymb} 
\usepackage{color}

\usepackage{multirow}
\usepackage{algorithm}
\usepackage{algorithmic}

\usepackage{array}

\begin{document}
\pagestyle{headings}

\title{End-to-End Multi-Object Tracking with Global Response Map} 



%
\author{Xingyu Wan \and
Jiakai Cao \and
Sanping Zhou \and
Jinjun Wang}
%
%
\institute{Institute of Artificial Intelligence and Robotics, \\Xi'an Jiaotong University, Xi'an, Shaanxi, 710049, China\\
\email{xingyuwan@stu.xjtu.edu.cn}\\
}


\maketitle

\begin{abstract}
Most existing Multi-Object Tracking (MOT) approaches follow the Tracking-by-Detection paradigm and the data association framework where objects are firstly detected and then associated. Although deep-learning based method can noticeably improve the object detection performance and also provide good appearance features for cross-frame association, the framework is not completely end-to-end, and therefore the computation is huge while the performance is limited. To address the problem, we present a completely end-to-end approach that takes image-sequence/video as input and outputs directly the located and tracked objects of learned types. Specifically, with our introduced multi-object representation strategy, a global response map can be accurately generated over frames, from which the trajectory of each tracked object can be easily picked up, just like how a detector inputs an image and outputs the bounding boxes of each detected object. The proposed model is fast and accurate. Experimental results based on the MOT16 and MOT17 benchmarks show that our proposed on-line tracker achieved state-of-the-art performance on several tracking metrics.

\end{abstract}

\section{Introduction}
Multi-Object Tracking (MOT) aims to use image measurements and predictive dynamic models to consistently estimate the states of multiple objects over discrete time steps corresponding to video frames. The major challenges of MOT are to continuously and effectively model the vast variety of objects with high uncertainty in arbitrary scenarios, caused by occlusions, illumination variations, motion blur, false alarm, etc~\cite{wan2018multi}. There are three key issues that a MOT framework should handle: 1) Modeling the dynamic motion of multiple objects; 2) Handling the entering/exiting of objects into/from the scene; 3) Robustness against occlusion and appearance/background variations. Single object tracking~\cite{held2016learning} focus on 1) and 3) but simply applying multiple single object trackers for the MOT task usually gives very limited performance due to 2).

With the significant progress in object detection, tracking-by-detection framework~\cite{nam2016learning} has become a leading paradigm whereby the detection results of objects are represented as bounding boxes and available in a video sequence as prior information. MOT is then casted as a problem of data association where the objective is to connect detection outputs into trajectories across video frames using suitable measurements. The performance of these approaches largely depends on two key factors: Firstly the quality of detection results, where if the detection is missing or inaccurate at a single frame, or when occlusion occurs, the target state is then hard to estimate, and the target identity is prone to be lost; Secondly the data association model, where to achieve robust association across frames given the dynamic of objects, many works~\cite{milan2013continuous,pirsiavash2011globally,son2017multi,tang2016multi} conduct MOT in an off-line fashion with iterative solver~\cite{zhang2015multi} in order to make use of detection from both past and future, but is therefore time consuming and sensitive to the quality of appearance feature for association, not to mention scenarios where on-line processing is required.

More recently, many works have been proposed to utilize deep learning techniques to train Convolutional Neural Networks (CNNs)~\cite{wang2015transferring,hong2015online} from large scale datasets to obtain rich feature representations. These models have significantly improved the object detection performance and the quality of appearance feature. Many MOT approaches~\cite{yang2012multi,sanchez2016online,wojke2017simple} have adopted Deep Neural Networks (DNNs) for feature representation learning and feature metric learning for data association. They usually establish a robust motion model to predict the motion variations of targets and introduce a well trained appearance model to extract deep feature from region of interest (ROI) for image patches, and finally some similarity distances are adopted to measure the affinity of two ROIs for pair-wise association. Furthermore, aiming to learn a robust metric for feature representation, several works~\cite{pellegrini2009you,scovanner2009learning,sadeghian2017tracking} take multiple features of objects in the scene by incorporating a myriad of components such as motion, appearance, interaction, etc. Some works~\cite{wan2018online,kim2018multi} even consider to combine temporal components to analyze long-term variation by using Long Short-Term Memory (LSTM). Since these methods are still based on disjoint detection/association steps, the computation is huge, and the performance is limited without end-to-end (i.e., from image-sequence/video to trajectory) capacity. There are works that attempt end-to-end training for the tracking-by-detection and association framework~\cite{zhang2020multiple,zhang2019frame}, but these approaches do not change the non-end-to-end nature of the MOT framework.

In this paper, we introduce a true end-to-end framework for MOT. The challenge is finding a suitable representation that is capable to handle both issues 1), 2) and 3) in an on-line manner. Our idea is to employ a modified object salience model to generate a global response map to locate the presence of multiple objects, such that for issue 1), the motion of each object is implicitly modeled, and for 2) and 3), minor occlusions/entering/exting within the window of frames can be robustly handled. The global response map has multiple channels where each channel models the response for different attributes to define the state space of all trajectories, such as ``presence of object'', ``$x$/$y$'', ``$\Delta x/\Delta y$'', as well as any additional attributes in the future, such that 1) the object detection step is only implicitly modeled in the tracking process, and the spatial information of a target is no longer a bounding box region of interest, but a Gaussian-like distribution from $0$ to $1$; and 2) the inference process does not need complicated assignment process but just a simple linking step to extract the multiple trajectories. In this paper, we applied a logical inference approach to estimate the actual state of target response based on the sequence of global response maps. Conceptually speaking, the sub-module for extracting each attribute of a state is similar to an ad-hoc network. For example, the attributes of ``presence of object'' and ``$x$/$y$'' are from a sub-module similar to an object locating network, the attribute of ``$\Delta x/\Delta y$'' is from a sub-module similar to an optical flow extraction network, etc. This can be further extended to include $width$, $height$, $orientation$, $depth$, etc by adding suitable sub-modules in the future. The most important part is that, we are able to integrate all these sub-modules into one end-to-end network for MOT in one feed-forward step without any exhaustive cropping and iterating. The overall framework is illustrated in Fig.~\ref{fig:framework}, and the main contributions of this work can be summarized as follows:

\begin{figure}[t]
\centering
\includegraphics[height=5.0cm]{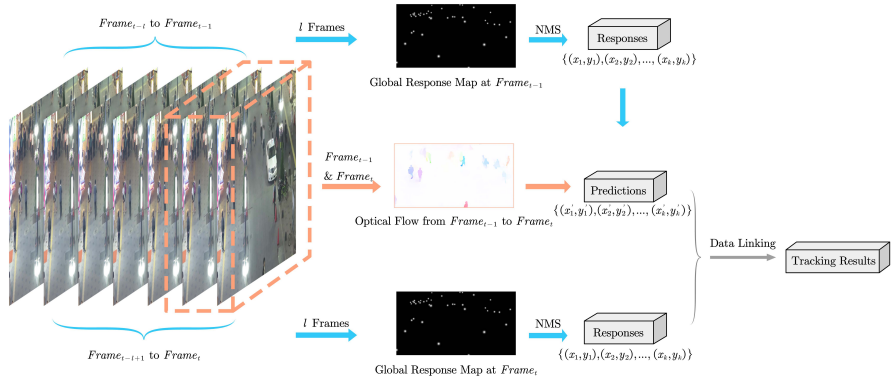}
\caption{The framework of our proposed MOT paradigm. This end-to-end framework is composed of several modules, which are object locating using global response map, motion displacement regression, and data linking using global assignment. The global response map is designed to extract the attributes of ``presence of object'' and ``$x/y$'', and motion displacement regressor is designed to retrieve the attribute of ``$\Delta x/\Delta y$''}
\label{fig:framework}
\end{figure}

1. We propose a novel representation schema and network structure to perform end-to-end MOT of learned object types. Significantly different from existing tracking-by-detection and data association based approaches, our proposed method takes image-sequence/video as input and generates the trajectories of multiple objects in a true end-to-end fashion, where multiple attributes that define the state space of each trajectory are obtainable from the global response map generated by our model.

2. The proposed network includes a sub-network that operates like an image-sequence/video-based object locator and is capable of handling the occlusion problem. From within the defined time window, the module can still maintain a positive response even when target is occluded, and thus significantly reducing the false negatives.

3. The proposed network also includes a sub-network that operates like an optical flow extraction network with a motion displacement regressor for estimating the motion dynamics. The module also helps solving the uncertain assignment problem in one single forward propagation.

4. The proposed multi-object tracking network is complete end-to-end without any detection/appearance priors. The experimental results show that our tracker achieves superior performance over the state-of-the-art approaches on public benchmarks.

\section{Related Works}
\subsubsection{Detector-based Tracking.}
Owe to the galloping progresses of object detection techniques such as Faster R-CNN~\cite{ren2015faster} and SDP~\cite{yang2016exploit}, given these detection results as intialization/priors, MOT task can be conducted within tracking-by-detection paradigm~\cite{nam2016learning} where the objective is to connect detection outputs into trajectories across video frames using reasonable measurements, which therefore casts the MOT problem as global data association. Traditional data association techniques including the Multiple Hypothesis Tracker (MHT)~\cite{reid1979algorithm} and the Joint Probabilistic Data Association Filter (JPDAF)~\cite{fortmann1983sonar} aim to establish sophisticated models to capture the combinatorial complexity on a frame-by-frame basis. Both methods got improved later in conjunction with better appearance model~\cite{kim2015multiple} or more efficient approximation~\cite{hamid2015joint}. Aiming at global optimization with simplified models, the flow network formulations~\cite{zhang2008global,pirsiavash2011globally,wang2015tracking,dehghan2015target} and probabilistic graphical models~\cite{yang2012online,yang2012multi,andriyenko2012discrete,milan2013detection} are considered, along with shortest-path, min-cost algorithms or even graph multi-cut formulations~\cite{tang2017multiple}. Most existing detector-based trackers highly rely on the quality of detection results, and to handle imperfect detections, several works~\cite{milan2013continuous,pirsiavash2011globally,son2017multi,tang2016multi}  conduct MOT in off-line fashion to handle ambiguous tracking results for a robust tracking performance. Due to their off-line nature with low processing speed, they are not applicable to real-time vision tasks.
Compare to these methods, our proposed approach runs in an on-line manner without being bounded to specific object detection techniques and does not require complicated data association step.

\subsubsection{Deep Metric Learning for MOT.}
Learning effective feature representation with corresponding similarity measure plays a central role in data association. Metric learning based on DNNs for object appearance representation and computation of the affinity between measurements has become a popular trend. Various trackers~\cite{pellegrini2009you,scovanner2009learning,sadeghian2017tracking,pellegrini2010improving,yamaguchi2011you,robicquet2016learning} model different features of objects by incorporating a myriad of components such as motion, appearance, interaction, social behavior, etc. Leal-Taixe et al.~\cite{leal2016learning} adopt a Siamese CNN to learn local features from both RGB images and optical flow maps. Robicquet et al.\cite{robicquet2016learning} introduced social sensitivity to describe the interaction between two targets and use this definition to help the data association step. Later on, inspired by the success of Recurrent Neural Networks (RNNs) and their application to language modeling~\cite{vinyals2015show}, several works have been trying to learn an end-to-end representation for state estimation utilizing RNNs~\cite{sadeghian2017tracking,milan2017online}. Sadeghian et al.~\cite{sadeghian2017tracking} proposed an off-line metric learning framework using a hierarchical RNN to encode long-term temporal dependencies across multiple cues, i.e., appearance, motion and interaction. Milan et al.~\cite{milan2017online} presented an on-line RNN-based approach for multiple people tracking which is capable of performing prediction, data association and state update within a unified network structure. Followed by these works, ~\cite{wan2018online} extended the research of RNN-based methods and leveraged the power of Long Short-Term Memory (LSTM) for learning a discriminative model of object trajectory by integrating dynamic features both in temporal and spatial. For the on-line MOT task, these methods may not perform well when heavy occlusion or mis-detection downgrade the robustness of appearance model. Differently, in our work, the occlusion problem is well handled by considering a window of frames to locate objects with an introduced logical inference methodology, without explicit appearance feature for metric learning.

\section{Our Proposed MOT Algorithm}
Traditional tracking-by-detection algorithm takes object detection and tracking as two separate tasks and adopts different CNNs respectively or applies cascade one. Within this procedure, different models and loss functions for different tasks are always needed, and therefore hard to achieve end-to-end training/inference. Aiming to integrating better DNNs-based detector into the visual tracking task, here we introduce an end-to-end MOT framework. Fig.~\ref{fig:framework} illustrates the proposed MOT framework, where we take consecutive image frames as network inputs, after learning a reasonable response map to locate interested targets globally, the target location is retrieved from response map using local NMS, and then we regress the motion displacements for these targets from a frame-wise optical-flow-like offset, after that we conduct global assignment between predictions and observations.

Our proposed MOT algorithm is organized as follows. The first section talks about the object locating sub-network, the next section is the motion displacement regression sub-network, and the final section is data linking strategy using global assignment.
\subsection{Object Locating using Global Response Map}
\begin{figure}[t]
\centering
\includegraphics[width=10.0cm]{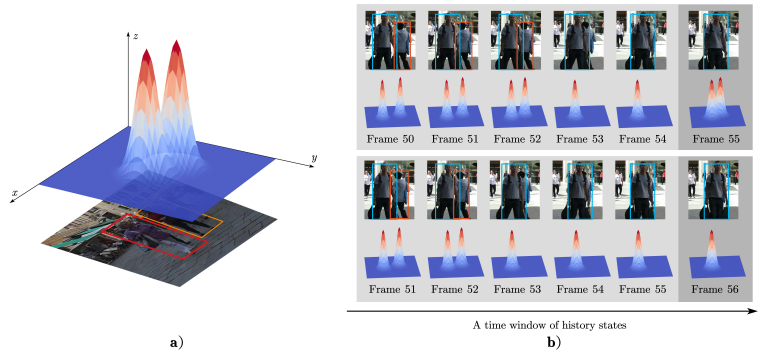}
\caption{Ground truth response maps, in which: $\textbf{(a)}$ An illustration of local response map viewed in 3D. $\textbf{(b)}$ An exemplary of logical inference methodology for target response estimation, row 1 and row 2 indicate the case 1 while row 3 and row 4 indicate the case 2. The actual states of two observed objects(annotated by blue and orange bounding boxes) at current frame (frame 55 for case 1, and frame 56 for case2) are estimated by a time window of history states}
\label{fig:logical}
\end{figure}

The goal of tracking is to consistently maintain the estimation of object states over discrete time step. In this specific computer vision task, using a well trained class-specific detector to filter out all the regions of interest over the image frame may not be necessary. Here we propose a simpler and more efficient way to locate objects for MOT. For all the targets to be tracked at each time step, we take them all as foreground objects and represent them as Gaussian-like distributions from $0$ to $1$ with peak value at their center points on a saliency map.
As shown in Fig.~\ref{fig:logical}$(a)$, each Gaussian-like distribution represents a foreground object to be tracked, the $x$, $y$ coordinates correspond to the object spatial location, and $z$ is a value from $0$ to $1$ corresponds to the actual status of object at current time step. The radius $r$ and sigma $\sigma$ of each distribution are defined as follow,
\begin{equation}
\label{eq:gaussian}
  r = \min_{i=1}^{3}|\frac{a_{i}+\sqrt{a_{i}^{2} - b_{i}}}{2}|, \qquad
  \sigma = \frac{r}{3}.
\end{equation}
where
\begin{align}
  a_{i} =
    \begin{cases}
      h+w, \ i=1 \\
      2\times(h+w), \ i=2 \\
      -2\times(h+w), \ i=3.
    \end{cases}
\end{align}
\begin{align}
  b_{i} =
    \begin{cases}
      4\times\frac{h \times w \times(1-\alpha)}{1+\alpha}, \ i=1 \\
      16\times h\times w\times(1-\alpha), \ i=2 \\
      16\times h\times w\times\alpha\times(\alpha-1), \ i=3.
    \end{cases}
\end{align}
Here $h$ and $w$ denote the height and width of target bounding box obtained from the ground-truth of training data, and $\alpha$ is an invariant parameter we set to 0.7 in this work. Given a bounding box scale $w$ and $h$, we first compute three different radius $\{r_{i},i=1,2,3\}$, and we adopt the minimum value as the radius $r$ of our Gaussian kernel, and the sigma $\sigma$ is set to $\frac{r}{3}$ accordingly. In this way, our Gaussian-like distribution has a positive correlation with target size $w$ and $h$.
\subsubsection{Response Map Learning Network.}
The above representation as global response map is able to describe the object spatial location and actual state at the same time. To learn such representation, our tracking algorithm employs a HED-based~\cite{xie2015holistically} saliency detection network modified from ~\cite{hou2017deeply}. Specifically, our object locating sub-network is an Auto-Encoder, which takes a time window of frames with a length $l$ as inputs, and outputs a single channel response map after a $sigmoid$ function. We adopt the short connection strategy as ~\cite{hou2017deeply} but remove the fusion layer, and we compute the average value of $1,2,3,6$ side outputs as our network output before activation. Given a training sequence $X_{l} = \{I^{t-l},...,I^{t}, I^{i} \in \mathbb{R}^{3*h*w}, i=t-l,...,t\}$, and label response map $Y \in \mathbb{R}^{1*h*w}= \{y_{j}, j=1,...,|Y|\}$, the standard cross entropy loss function for our network is given by
\begin{align}
  L(X_{l}, Y) &= -\sum_{j=1}^{|Y|}(y_{j}log\mathbf{P}(y_{j}=1|X_{l}) + (1-y_{j})log\mathbf{P}(y_{j}=0|X_{l})),
\label{eq:loss}
\end{align}
where $\mathbf{P}(y_{j}=1|X_{l})$ denotes the probability of the activation value at location j, and label $Y$ is obtained using the following logical inference methodology from the ground-truth of training data.
\subsubsection{Logical Inference Methodology for Handling Occlusion.}
When a target being tracked is unseen at a particular time step, that does not mean this target actually leaves the surveillance scene.
In this work, we argue that the actual state of object presence in visual tracking scenes should be distinguished from image-based detection results, and this estimation of actual state can be learned using history priors. As described above, we use a $0/1$ response value to represent the target actual state at each time step. This representation should be estimated using not only image at current frame, but also images from the past. Here we introduce a logical inference methodology for estimating response value of each target.
For target trajectories $\{T_{j}, j=1,2,...m\}$ from ground-truth, the response value $z_{j}^{t}$ of target actual state at frame $t$ is estimated upon a time window of past states $\{z_{j}^{i}, i=t-l,t-l+1,...,t-1\}$ with a length $l$. The specific estimation method is described as follow,
\begin{align}
z_{j}^{t} =
    \begin{cases}
      1, & \mbox{if } \; z_{j}^{t-1}=1 \;\mbox{or }\; \frac{\sum_{i=t-l}^{i=t-1}z_{j}^{i}}{l}>=\beta \\
      0, & \mbox{otherwise}.
    \end{cases}
\end{align}
Here $\beta$ is a constant describing the proportion of positive states within the time window. In general, if a target appeared to be existing during most of the past time, we take it still exist at current time step even when we get a negative result from image-based detector. Conversely, if a target remained negative during most of the past time, we consider it actually leave the scene and take the positive detection of current time step as a false alarm. In addition, if a target has a positive response at last time step, i.e., $z_{j}^{t-1}=1$, we set $z_{j}^{t}=1$ accordingly no matter what the history states are. At training phase, this estimation strategy is employed to generate positive/negative samples from training data. By using this strategy, we can consistently maintain the correct positive responses for targets even when suffering occlusions.

\subsection{Motion Displacement Regression}
\begin{figure}[t]
\centering
\includegraphics[width=10.0cm]{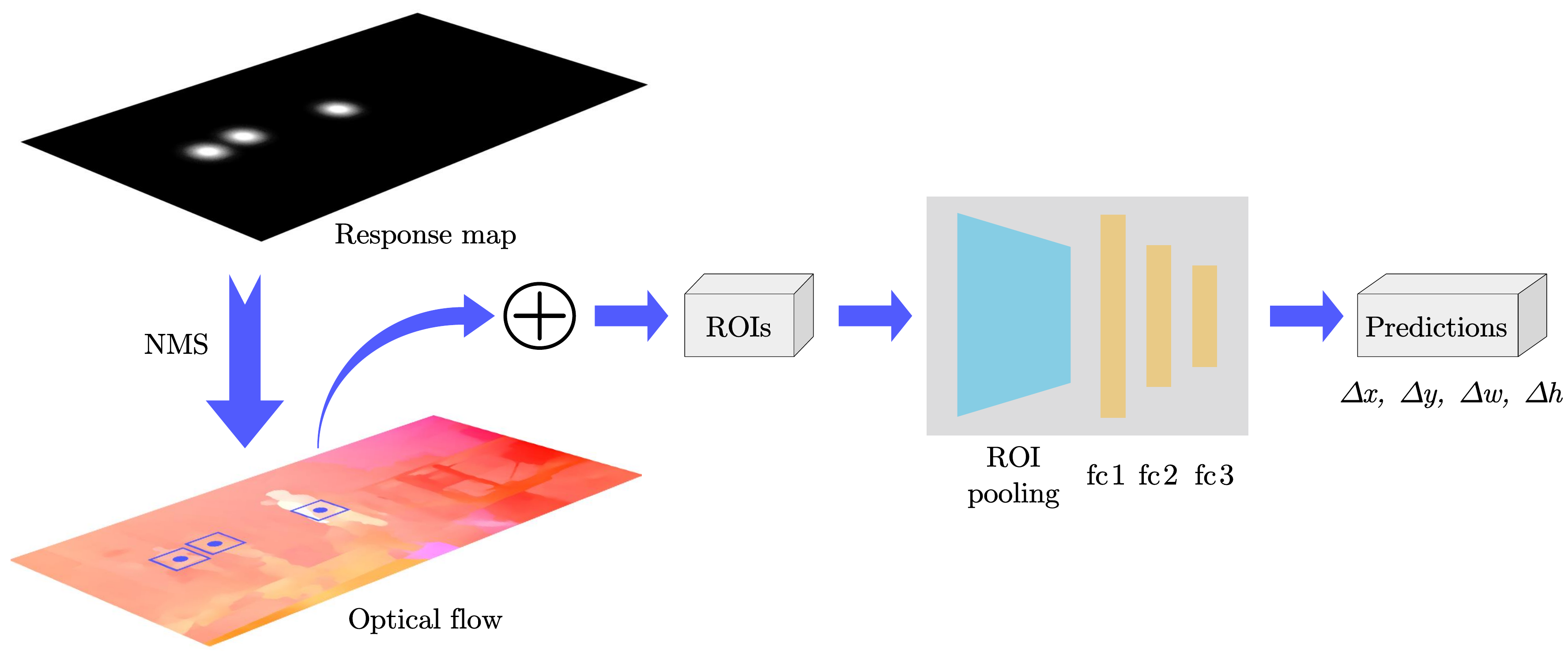}
\caption{The network structure of motion displacement regression. The inputs of this network are the center point distributions of located objects retrieved from global response map after NMS. The outputs are predicted motion displacements of all positive responses}
\label{fig:optical}
\end{figure}
To obtain the predicted motion displacements of all located targets, most popular approaches are to iteratively crop the image patch of each region of interest(ROI) to learn a regressed motion prediction. These approaches are time consuming and hard to achieve end-to-end training/inference.
In our proposed MOT algorithm, the motion dynamics are estimated from a motion displacement regression sub-network using a frame-wise optical-flow-like offset.
As proposed in ~\cite{ilg2017flownet}, we adopt FlowNet2 for frame-wise optical flow estimation. Given two adjacent frames $I^{t-1}$ and $I^{t}$, the optical flow estimation from frame $t-1$ to frame $t$ can be derived as $W^{t-1} = \sum_{i}(u_{i},v_{i}), W^{t-1} \in \mathbb{R}^{2*h*w}$, where $i$ denotes each pixel on the flow. After deriving the pixel-wise displacements from optical flow, we introduce a regression network to learn the predicted motion displacements globally for all responses with Gaussian-like distributions. As shown in Fig.~\ref{fig:optical}, we first conduct local NMS with a kernel size $s$ on response map $Z^{t}$ at frame $t$ with a threshold value of response $Score$ to filter out the top $k$ positive responses and retrieve their center point locations. For each retrieved response distribution with a center point $(cx,cy)$ and fixed kernel size $r_{z}$, we take it as a region of interest (ROI), and sample all the displacements of ROIs from optical flow at the same time to obtain a concatenated featuremap $F^{t-1} \in \mathbb{R}^{k*2*r_{z}*r_{z}}$ for regression. The regression network is composed of a ROI pooling~\cite{girshick2015fast} layer and several fully-connected layers. This network structure is designed to learn one accurate displacement value of response point from a ROI. The network output $D^{t} = \{d_{j}=(\Delta cx, \Delta cy, \Delta w, \Delta h), j=1,...,k\}$ is a movement displacement vector of all response points from frame $t-1$ to frame $t$. Given the ground-truth $G^{t}$ and network output $D^{t}$, the loss function of our regression network is defined as
\begin{align}
smooth_{L_{1}}(G^{t},D^{t}) =
    \begin{cases}
      0.5(x)^{2}, & \mbox{if } |x|<1 \\
      |x|-0.5, & \mbox{otherwise}.
    \end{cases}
\end{align}
where $x$ denotes the L1 loss between $G^{t}$ and $D^{t}$.
Incorporating ROI pooling into our regression network enables our tracker to estimate frame-wise motion displacements of all the observed targets in one single forward propagation without any cropping and network iteration.

\subsection{Data Linking Strategy}
At tracking phase, the global response map $Z^{t}$ and motion displacement estimation $D^{t}$ are already obtained using the proposed framework, then our tracker conduct global assignment between predictions and observations. We first obtain the predicted location $(cx^{'},cy^{'},w^{'},h^{'})$ by adding the regressed motion displacement $(\Delta cx, \Delta cy, \Delta w, \Delta h)$ to the previous target location. Followed by this, we compute the assignment cost matrix between predictions $D^{t}$ and observations $Z^{t}$ with the intersection-over-union (IOU) distance defined as Eq.~\ref{eq:iou}, then we solve the assignment problem optimally using the Hungarian algorithm~\cite{kuhn1955hungarian}.
\begin{align}
IOU(a,b) = \frac{Area(a) \cap Area(b)}{Area(a) \cup Area(b)}
\label{eq:iou}
\end{align}
Specifically, we first compute the IOU distance $IOU(D^{t},Z^{t})$ between each predicted location $d_{k}^{t} \in D^{t}$ with its nearest neighbourhood response $z_{k}^{t} \in Z^{t}$ at frame t. The nearest neighbourhood response is picked by solving the shortest path using center points distance of two distributions. After that, we pick the responses from observation space with the max IOU value higher than $IOU_{min}$ as the candidate, and compute the IOU distance between the response locations at frame $t-1$ and the candidate locations at frame $t$. Followed by this, the assignment problem leads to an optimal association between detections and candidates which can be solved by applying the Hungarian algorithm~\cite{kuhn1955hungarian} to maximize the sum of all IOUs at frame $t$.
After this global assignment approach, we then adopt a matching cascade strategy for all unmatched responses and tracks at current frame $t$ similar as ~\cite{wojke2017simple}. Specifically, we set a constant parameter $A_{max}$ denotes the maximum age. For each response $z_{k}^{t}$ not assigned to an existing track at current frame, we compute the IOU distance between this response with each terminated track whose last frame is within a time window from $t-1$ to $t-A_{max}$. This computation is iteratively conducted frame-by-frame until the IOU distance is higher than $IOU_{min}$, and we take this response $z_{k}^{t}$ to update the target state of corresponding terminated track as a match.
After this matching cascade, all responses not assigned to an existing track will be initialized as a new track, and all tracks without an assigned response will be terminated.

\section{Experiment}
\subsection{Experimental Details}
\begin{figure}[t]
\centering
\includegraphics[width=12.0cm]{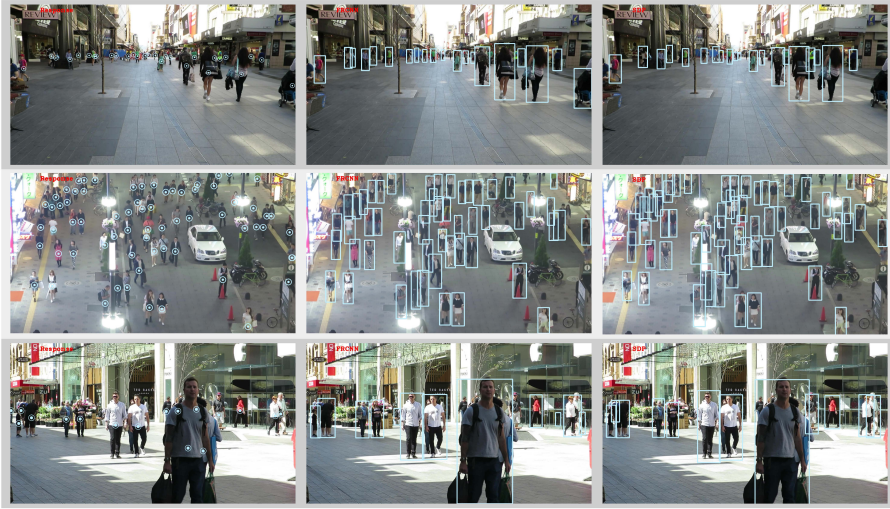}
\caption{Qualitative results of our object locating method (column 1) and two popular detectors (Faster-RCNN~\cite{ren2015faster}(column 2) and SDP~\cite{yang2016exploit}(column 3) respectively) on some challenging cases, best viewed in color. Our object response can still be located on the object even when occluded, which significantly helps the following data linking module to obtain a complete trajectory for the object}
\label{fig:detection}
\end{figure}
We report the performance of the proposed MOT algorithm on the MOT16~\cite{milan2016mot16} and MOT17~\cite{milan2016mot16} benchmark datasets. The provided public detection results of MOT16 dataset is DPM~\cite{felzenszwalb2009object}. The MOT17 dataset contains the same video sequences(7 fully annotated training sequences and 7 testing sequences) as MOT16 but with two more sets of public detection results from Faster-RCNN~\cite{ren2015faster} and SDP~\cite{yang2016exploit} respectively.
We implemented our framework in Python3.6 using PyTorch, with six cores of 2.4GHz Inter Core E5-2680 and a NIVIDIA GTX 1080 GPU.

For object locating, here we set the length $l$ of time window to $5$. The parameter $\beta$ of logical estimation methodology is set to $0.6$. At the training phase, we use the VGGNet~\cite{simonyan2014very} pre-trained on the ImageNet dataset~\cite{deng2009imagenet} as the shared base convolution blocks. The input of our network is a sequence of image frames which are resized to $512\times960$ for height and width. The total training epochs are $300$ and the learning rate is initialized to $1e-3$ and divided by $10$ every $100$ epochs.

For motion displacement regression, we trained our model on 7 MOT training sequences with provided ground-truth tracking results. Here we set local NMS kernel size $s$ to $3$ and the threshold value of response $Score$ to $0.05$ , the parameter $k$ for maximum number of positives to $60$, and the fixed kernel size $r_{z}$ of ROI to $20$. At the training phase, we use the FlowNet2~\cite{ilg2017flownet} pre-trained on MPI-Sintel~\cite{butler2012naturalistic} for extracting frame-wise optical flow. The input of our network for motion variation regression is a sequence of adjacent image frames which are resized to $1024\times1920$ for height and width. We set the initial learning rate to $1e-4$, the total training epoches to $500$ and divide the learning rate by $2$ at epoch $166$, $250$, $333$ and $416$ respectively. We use Adam~\cite{kingma2014adam} optimizer for both sub-networks. The parameters $IOU_{min}$ and $A_{max}$ for data linking are set to $0.7$ and $30$ respectively.
\subsection{Object Locating Performance}
In order to exam the validity of our proposed representation schema, we visualized the results of object locating sub-network as well as the provided detection results on some challenging sequences from MOT benchmark~\cite{milan2016mot16}. As shown in Fig.\ref{fig:detection}, compared with the official detection results from Faster-RCNN~(column 2) and SDP~(column 3) on challenging scenarios like small scales~(row 1), crowded~(row 2) and occluded~(row 3), our object locating method~(column 1) gives more accurate estimations centered on the foreground objects. Furthermore, at extremely occluded scenario~(row 3), benefit from logical inference on history states, our method can still give positive and accurate responses for those objects completely occluded at this frame, while detector-based trackers have to make a further analyze on such case by introducing more complicate tricks.
\begin{table}[t]
\begin{center}
\caption{Run-time comparison among our proposed object locater and several popular detectors}
\label{table:time1}
\scriptsize
\begin{tabular}{p{5.0cm}<{\centering} p{2.5cm}<{\centering} p{2.5cm}<{\centering}}
\hline\noalign{\smallskip}
Method & ms/frame & FPS\\
\hline
Fastest-DPM~\cite{yan2014fastest}  & 66 & 15\\
Fast-RCNN~\cite{girshick2015fast}  & 320 & 3\\
Faster-RCNN~\cite{ren2015faster}  & 198 & 5\\
YOLO~\cite{redmon2016you} & 22 & 45\\
SSD300~\cite{liu2016ssd} & 17 & 58\\
Response Map ($\textbf{Ours}$) & $\textbf{5}$ & $\textbf{200}$\\
\hline
\end{tabular}
\end{center}
\end{table}

In addition to the accuracy term, our object locating method runs much faster than DNN-based detectors. In Table~\ref{table:time1}, we compared the running time of our proposed object locating method with several popular detectors. Here we report the running times of classic DPM detector with a speeding up version~\cite{yan2014fastest}, two-stage CNN-based detectors Fast-RCNN of VGG-16 version~\cite{girshick2015fast} and Faster-RCNN with VGG-16 for both proposal and detection~\cite{ren2015faster}, one-stage CNN-based detectors YOLO using VGG-16~\cite{redmon2016you} and SSD with $300 \times 300$ image size~\cite{liu2016ssd}.
Our method were measured by computing the average run-time on 7 test sequences from MOT benchmark~\cite{milan2016mot16}. The FPS of our method for one feed-forward propagation is $40$ times faster than classic Faster-RCNN~\cite{ren2015faster} and about $3-4$ times faster than real-time detectors YOLO~\cite{redmon2016you} and SSD~\cite{liu2016ssd}.
These evaluation results confirm that object locating from the global response map is faster and more accurate. But more importantly, different from tracking-by-detection and association based methods, this object locater is utilized as a sub-network within our MOT algorithm in an end-to-end fashion.
\subsection{Tracking Performance on MOT Benchmark}
\subsubsection{Tracking as Response Points.}
\begin{figure}[t]
\centering
\includegraphics[height=4.0cm]{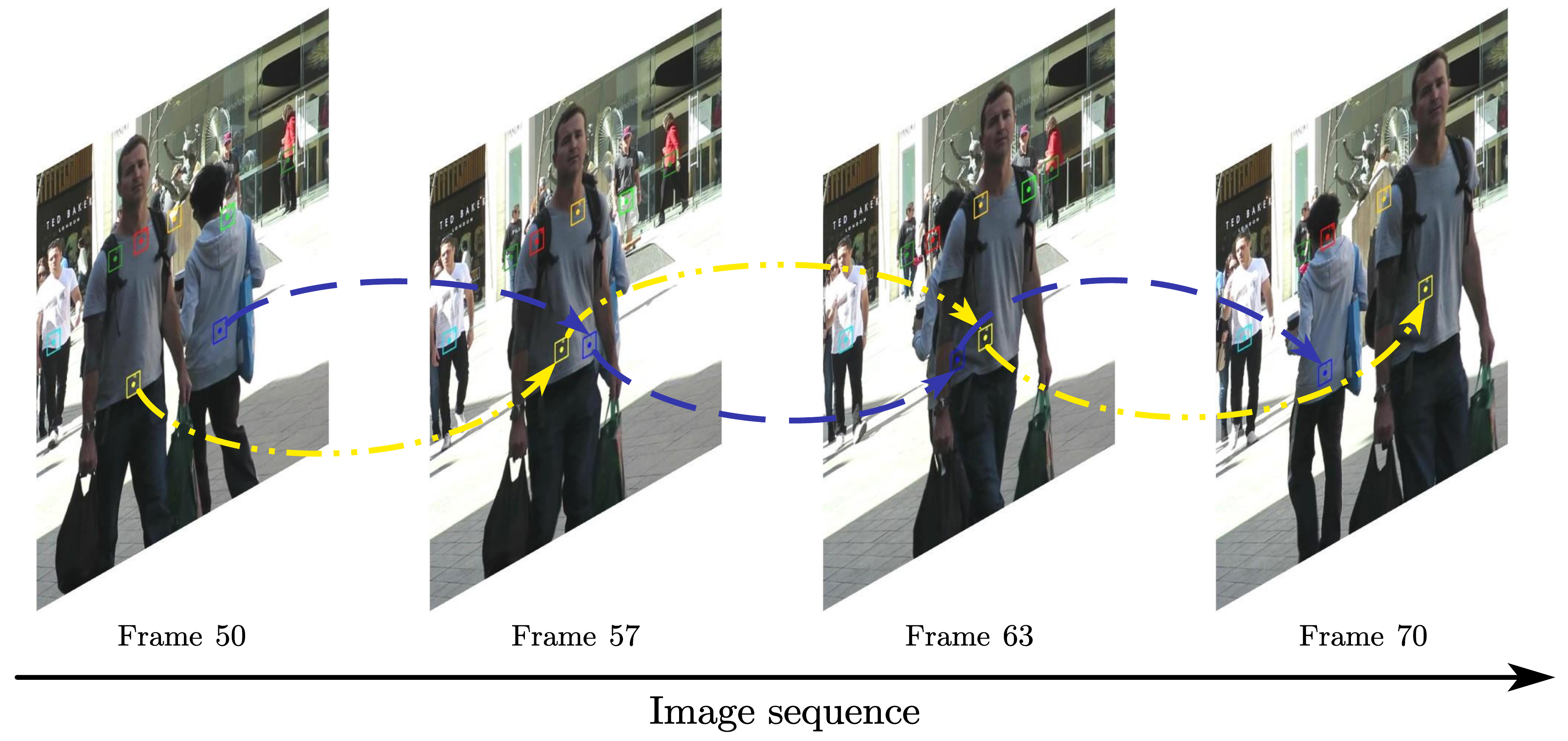}
\caption{An illustration of tracking results visualization using tracking as response points method on an extremely occluded scene.}
\label{fig:track}
\end{figure}

Different from traditional tracking-by-detection paradigm, our proposed MOT method operates without using any detection results. Hence, instead of using the provided bounding box coordinates, our tracker only takes raw image-sequence/video as inputs, and tracks on each foreground response distribution. The representation of this tracking response points includes a center point coordinates $(cx,cy)$ and a fixed kernel size $r_{z}$ for each target.
The exemplary visualization of our tracking as response points method on the MOT challenge dataset is shown in Fig.~\ref{fig:track}. Owe to the accurate center point locations retrieved from global response map and effective motion model, our tracker is rather robust for maintaining the identities of tracked targets during short-term occlusions.
\subsubsection{Comparison to State-of-the-Art Trackers.}
Since the existing evaluation metrics for MOT are designed for traditional tracking-by-detection paradigm where the tracking results are provided and expressed as bounding box coordinates, in order to compare the performance of our tracker with other tracking methods, we ran one additional step that maps our generated response points to the provided detection bounding boxes from MOT challenge benchmark~\cite{milan2016mot16} so that the required metrics can then be calculated. Specifically, we compute the cost matrix between response distributions and bounding box detections using center point distance. Then we make the global optimal assignment by using Hungarian algorithm~\cite{kuhn1955hungarian} same as section 3.3. We take the retrieved positive responses from object locating network as priors. For those detections from MOT dataset but not assigned to any response, we take them as false alarms. For the responses not assigned to any detection, we maintain the object location with an initialized bounding box scale, and this scale is adjusted by motion variation regression network during tracking process.
\begin{table}[t]
\begin{center}
\caption{Evaluation results on the MOT16 dataset}
\label{table:mot16}
\tiny
\begin{tabular}{ccccccccccccc}
\hline
Mode& Method & MOTA $\uparrow$ & MOTP $\uparrow$ & IDF1 $\uparrow$ & IDP $\uparrow$ & IDR $\uparrow$ & MT $\uparrow$ & ML $\downarrow$ & FP $\downarrow$ & FN $\downarrow$ & ID Sw. $\downarrow$ & Frag $\downarrow$\\
\hline
$\times$&STRN16~\cite{xu2019spatial}&48.5&73.7&53.9&72.8&42.8&17.0\%&34.9\%&9,038&84,178&747&2,919\\
$\circ$&LMP~\cite{tang2017multiple}&48.8&$\textbf{79.0}$&51.3&71.1&40.1&18.2\%&40.1\%&6,654&86,245&461&595\\
$\circ$&KCF16~\cite{chu2019online}&48.8&75.7&47.2&65.9&36.7&15.8\%&38.1\%&5,875&86,567&906&1,116\\
$\times$&AFN~\cite{shen2018tracklet}&49.0&78.0&48.2&64.3&38.6&19.1\%&35.7\%&9,508&82,506&899&1,383\\
$\times$&eTC~\cite{wang2019exploit}&49.2&75.5&56.1&$\textbf{75.9}$&44.5&17.3\%&40.3\%&8,400&83,702&606&882\\
$\circ$&LSST16~\cite{feng2019multi}&49.2&74.0&56.5&77.5&44.5&13.4\%&41.4\%&7,187&84,875&606&2,497\\
$\circ$&HCC~\cite{ma2018customized}&49.3&$\textbf{79.0}$&50.7&71.1&39.4&17.8\%&39.9\%&5,333&86,795&$\textbf{391}$&$\textbf{535}$\\
$\circ$&NOTA~\cite{chen2019aggregate}&49.8&74.5&55.3&75.3&43.7&17.9\%&37.7\%&7,248&83,614&614&1,372\\
$\circ$&Tracktor16~\cite{bergmann2019tracking}&54.4&78.2&52.5&71.3&41.6&19.0\%&36.9\%&$\textbf{3,280}$&79,149&682&1,480\\
$\circ$&$\textbf{Ours}$&$\textbf{62.0}$&73.6&$\textbf{63.8}$&70.5&$\textbf{58.3}$&$\textbf{37.7\%}$
&$\textbf{20.7\%}$&18,308&$\textbf{50,039}$&909&2,009\\

\hline
\end{tabular}
\end{center}
\end{table}

We evaluate our tracker on the test sets of both MOT16 and MOT17 benchmark. The evaluation is carried out according to the metrics used by the MOT benchmarks~\cite{bernardin2008evaluating,ristani2016performance,li2009learning}, which includes Multiple Object Tracking Accuracy (MOTA), Multiple Object Tracking Precision (MOTP),
ID F1 Score (IDF1, the ratio of correctly identified detections over the average number of ground-truth and computed detections), ID Precision (IDP), ID Recall (IDR), Mostly tracked targets (MT), Mostly lost targets (ML), the total number of false positives (FP), the total number of false negatives (FN), the total number of identity switches (ID Sw.), the total number of times a trajectory is fragmented (Frag).

Table~\ref{table:mot16} and Table~\ref{table:mot17} present the quantitative performance respectively, in comparison with some of existing best performing published trackers both online and offline. Evaluation metrics with $\uparrow$ means that higher scores denote better performance, while $\downarrow$ means the lower scores denote better performance. Mode with mark $\circ$ means an online method, while mark $\times$ means an offline method. From Table~\ref{table:mot16} and Table~\ref{table:mot17} we can see that the main evaluation metric MOTA of our proposed method surpasses all the other state-of-the-art trackers. Moreover, our tracker also achieves the best performance in IDF1, IDR, MT, ML and FN. Among these metrics, the highest MT and the lowest ML indicate the robustness of our tracker for maintaining the identity of tracked targets. Futhermore, the lowest FN confirms introducing object locating network to estimate the actual targets does make an effort for reducing the false negatives caused by image-based detector, and the occlusion problem is handled rather well.
\begin{table}[t]
\begin{center}
\caption{Evaluation results on the MOT17 dataset}
\label{table:mot17}
\tiny
\begin{tabular}{ccccccccccccc}
\hline
Mode& Method & MOTA $\uparrow$ & MOTP $\uparrow$ & IDF1 $\uparrow$ & IDP $\uparrow$ & IDR $\uparrow$ & MT $\uparrow$ & ML $\downarrow$ & FP $\downarrow$ & FN $\downarrow$ & ID Sw. $\downarrow$ & Frag $\downarrow$\\
\hline
$\times$&STRN17~\cite{xu2019spatial}&50.9&75.6&56.0&74.4&44.9&18.9\%&33.8\%&25,295&249,365&2,397&9,363\\
$\times$&jCC~\cite{keuper2018motion}&51.2&75.9&54.5&72.2&43.8&20.9\%&37.0\%&25,937&247,822&1,802&2,984\\
$\circ$&NOTA~\cite{chen2019aggregate}&51.3&76.7&54.5&73.5&43.2&17.1\%&35.4\%&20,148&252,531&2,285&5,798\\
$\circ$&FWT~\cite{henschel2018fusion}&51.3&77.0&47.6&63.2&38.1&21.4\%&35.2\%&24,101&247,921&2,648&4,279\\
$\times$&AFN17~\cite{shen2018tracklet}&51.5&77.6&46.9&62.6&37.5&20.6\%&35.5\%&22,391&248,420&2,593&4,308\\
$\times$&eHAF17~\cite{sheng2018heterogeneous}&51.8&77.0&54.7&70.2&44.8&23.4\%&37.9\%&33,212&236,772&1,834&$\textbf{2,739}$\\
$\times$&eTC17~\cite{wang2019exploit}&51.9&76.3&58.1&73.7&48.0&23.1\%&35.5\%&36,164&232,783&2,288&3,071\\
$\times$&FAMNet~\cite{chu2019famnet}&52.0&76.5&48.7&66.7&38.4&19.1\%&33.3\%&14,138&253,616&3,072&5,318\\
$\circ$&JBNOT~\cite{henschel2019multiple}&52.6&77.1&50.8&64.8&41.7&19.7\%&35.8\%&31,572&232,659&3,050&3,792\\
$\circ$&LSST17~\cite{feng2019multi}&54.7&75.9&62.3&$\textbf{79.7}$&51.1&20.4\%&40.1\%&26,091&228,434&$\textbf{1,243}$&3,726\\
$\circ$&Tracktorv2~\cite{bergmann2019tracking}&56.3&$\textbf{78.8}$&55.1&73.6&44.1&21.1\%&35.3\%&$\textbf{8,866}$&235,449&1,987&3,763\\
$\circ$&$\textbf{Ours}$&$\textbf{61.2}$&74.8&$\textbf{63.2}$&70.4&$\textbf{57.3}$&$\textbf{36.7\%}$
&$\textbf{22.0\%}$&55,168&$\textbf{159,986}$&3,589&7,640\\

\hline
\end{tabular}
\end{center}
\end{table}
\subsubsection{Run-Time Efficiency.}
We investigate the run-time efficiencies of each module in our proposed MOT method. At online inference phase, the run-times of each component tested on MOT benchmark dataset~\cite{milan2016mot16} is listed in Table~\ref{table:time2}. Our tracking as response points method outputs trajectories for all responses in one feed-forward propagation, and the run-time (ms per image) of our tracker is $162.5ms$($6$ FPS). This is noticeably faster than many ``real-time'' and online methods ~\cite{fang2018recurrent,yu2016poi,mahmoudi2019multi,zhou2018online,wojke2017simple} where ``tracking'' is conducted after a non-negligible detection step, such as those from the MOT benchmark~\cite{milan2016mot16} based on Faster-RCNN~\cite{ren2015faster} detectors (as shown in Table~\ref{table:time3}).
\begin{table}[t]
\begin{center}
\caption{Timing of each components of our proposed algorithm}
\label{table:time2}
\scriptsize
\begin{tabular}{p{5.0cm}<{\centering} p{2.5cm}<{\centering} p{2.5cm}<{\centering}}
\hline\noalign{\smallskip}
Component & ms/frame & FPS\\
\hline
Object locating~(including NMS)  & 108 & 9\\
Motion displacement regression & 7.5 & 133\\
Data association & 47 & 21\\
Total & 162.5 & 6\\
\hline
\end{tabular}
\end{center}
\end{table}

\begin{table}[t]
\begin{center}
\caption{Run-time(ms/frame) comparison on MOT challenge benchmark}
\label{table:time3}
\scriptsize
\begin{tabular}{p{3.0cm}<{\centering} p{2.0cm}<{\centering} p{2.0cm}<{\centering} p{2.0cm}<{\centering} p{2.0cm}<{\centering}}
\hline\noalign{\smallskip}
Method & Detection & Tracking & Total & FPS\\
\hline
RAR16wVGG~\cite{fang2018recurrent} & $198$ & $\textgreater$600 & $\textgreater$798 &$\textless$1.5\\
POI~\cite{yu2016poi}  & $198$ & $\textgreater$160 & $\textgreater$358 &$\textless$3\\
CNNMTT~\cite{mahmoudi2019multi} & $198$ & $\textgreater$150 & $\textgreater$348 &$\textless$3\\
TAP~\cite{zhou2018online}  & $198$ &$\textgreater$120 & $\textgreater$318 &$\textless$3.5\\
Deep SORT~\cite{wojke2017simple} & $198$ & $\textgreater$25 & $\textgreater$223 &$\textless$4.5\\
$\textbf{ours}$ & $\textbf{0}$ & $\textbf{162.5}$ & $\textbf{162.5}$ &$\textbf{6}$\\
\hline
\end{tabular}
\end{center}
\end{table}
\section{Conclusions}
In this work, we introduce a novel object representation schema and a new network model to support end-to-end on-line MOT. The proposed approach is significantly different from existing tracking-by-detection and association based methods where the object detection step is only implicitly conducted by a more efficient object locating network, while other attributes of a tracked object, such as $x/y$ and $\Delta x/\Delta y$, can be extracted by the corresponding sub-networks (the object locating sub-network and the motion displacement regression sub-network respectively) in one feed-forward propagation. The complete model is capable of describing the dynamic motions of multiple objects and can handle the entering/exiting/occlusion of objects robustly. The network generates a global response map as the intermediate output from which the trajectory of each object can be obtained. The proposed method is fast and accurate, and our evaluation based on the MOT benchmark show that the proposed tracker significantly outperforms many other state-of-the-art methods. We believe that such simple and effective algorithm can provide a new inspiration for the MOT community. Our extension to extract other attributes in a more complete motion tracking state space, such as width $w$ and height $h$ of the object, $\Delta w$ and $\Delta h$, as well as other attributes including $orientation$, $depth$, etc, is on-going.

%
%
\bibliographystyle{splncs04}
\bibliography{egbib}
\end{document}